# Impact of Adversarial Attacks on Deep Learning Model Explainability

Gazi Nazia Nur, Mohammad Ahnaf Sadat

Iowa State University

Email address: sadat@iastate.edu

ABSTRACT

In this paper, we investigate the impact of adversarial attacks on the explainability of deep learning models, which are commonly criticized for their black-box nature despite their capacity for autonomous feature extraction. This black-box nature can affect the perceived trustworthiness of these models. To address this, explainability techniques such as GradCAM, SmoothGrad, and LIME have been developed to clarify model decision-making processes. Our research focuses on the robustness of these explanations when models are subjected to adversarial attacks, specifically those involving subtle image perturbations that are imperceptible to humans but can significantly mislead models. For this, we utilize attack methods like the Fast Gradient Sign Method (FGSM) and the Basic Iterative Method (BIM) and observe their effects on model accuracy and explanations. The results reveal a substantial decline in model accuracy, with accuracies dropping from 89.94% to 58.73% and 45.50% under FGSM and BIM attacks, respectively. Despite these declines in accuracy, the explanation of the models measured by metrics such as Intersection over Union (IoU) and Root Mean Square Error (RMSE) shows negligible changes, suggesting that these metrics may not be sensitive enough to detect the presence of adversarial perturbations.

# 1    Introduction

Computational perception capabilities have impacted our daily lives and industries in a revolutionary way. These capabilities include but are not limited to image recognition, speech recognition, augmented reality, sentiment analysis, and natural language processing. In this paper, however, our focus is solely on image recognition.

The influence of image recognition, powered by computer vision, is profoundly evident in manufacturing. Automated inspection systems significantly enhance product quality by identifying defects in components. Similarly, automated object recognition systems modernize inventory management, reducing manual labor and boosting profitability and productivity. Furthermore, the retail sector also benefits from advancements in customer experience, notably through implementing self-checkout systems.

Before the emergence of deep learning, computer vision predominantly utilized conventional techniques and algorithms. These methods frequently depended on handcrafted features and rules-based systems for image analysis. Common practices included feature extraction techniques like edge detection to discern important image patterns and structures. These approaches faced difficulties with real-world data's complexity and diversity, relying on handcrafted features and having a restricted ability to learn from vast datasets. Despite such hurdles, these approaches are still applied in the industrial automation, surveillance, and medical imaging sectors.

Although our goal in this paper is not to compare deep learning models with hand-crafted algorithms, we still present how hand-crafted algorithms excel with their explainability to demonstrate what explainability means. As an example of explainability, let us consider the problem of receipt classification in the MATLAB image processing onramp (refer to Figures 1 and 2). In this case, we check if an image is a receipt by analyzing the oscillations of row sums ( row sums are the totals of black pixel values in each row, and observing how these sums change from one row to the next). Specifically, if the oscillation reveals fewer than nine minima, we classify the image as non-receipt. Conversely, if there are nine or more minima, it is identified as a receipt. This method enabled us to categorize two distinct figures effectively. Furthermore, this approach offers a clear explanation. For example if an image is not recognized as a receipt, in that case, it lacks at least nine minima in its row sum oscillation, providing clear insight into the classification and its rationale.

However, this method comes with certain drawbacks, including the need for specific domain knowledge, a labor-intensive design process, and the challenge of creating rules for large datasets, which can lead to reduced accuracy.

These challenges have been effectively addressed by deep learning, which has revolutionized computer vision and image processing. Instead of using manually created features and rules like old methods, deep learning, especially through Convolutional Neural Networks (CNNs), learns directly from the images themselves. This means we do not develop the algorithm manually. Moreover, deep learning can work well with massive amounts of data. That is why it has been widely adopted in the agriculture, manufacturing, medical, and autonomous vehicle industries.

Although deep learning models are extensively capable of capturing complex relationships within data, their inherent model complexity often makes them appear as "black boxes" (Rudin, 2019), obscuring the reasons behind their outputs. This can lead to a lack of trust in machine learning models, especially in critical sectors such as food and pharmaceuticals. Addressing the challenge of explainability of deep learning models has emerged as a pivotal area of research. Explainable artificial intelligence (XAI) deals with this issue. XAI offers insights that clarify the reason behind the deep learning model's prediction and help build user trust. In the context of image classification, XAI can pinpoint the exact pixels influencing the model's decisions. In this paper, we will refer to these pixels as the explanation of the predictions made by the deep learning model.

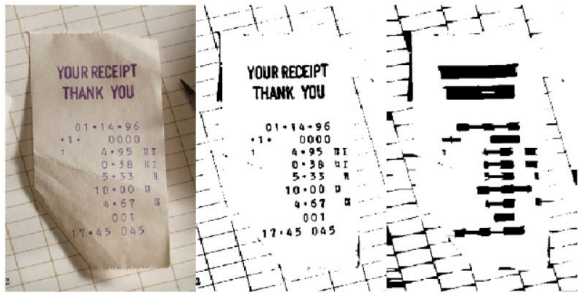

Figure 1: Positive receipt classification in MATLAB Image Processing Onramp (mathworks.com, n.d.)

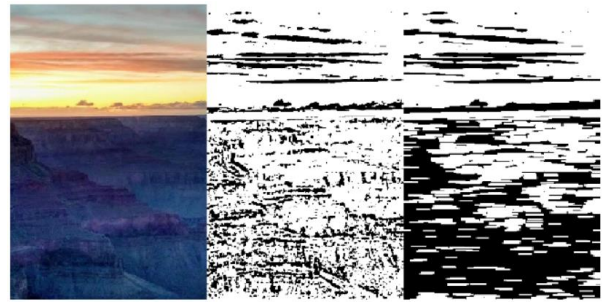

Figure 2: Negative receipt classification in MATLAB Image Processing Onramp (mathworks.com, n.d.)

In addition to explainability, another emerging research area in deep learning is adversarial attacks. Adversarial attacks exploit the complex decision-making processes of deep learning models by introducing a minimal amount of noise to an original image. This noise is usually imperceptible to humans but sufficient to mislead the models. Such decision-time adversarial attacks add a very small amount of noise to the original image. This noise is so slight that humans cannot notice it, but it is enough to confuse the deep learning models.

Adversarial attacks further complicate trust issues with deep learning. Understanding how these attacks alter the explanations provided by models is crucial. In this paper, we aim the following:

1. How do the explanations of a deep learning model for an image classification task change in the presence of adversarial attacks?

2. How effective are different explanation techniques when faced with adversarial conditions?

This study aims to enhance our understanding of how adversarial attacks affect the explainability of deep learning model predictions and seeks to establish a foundation for benchmarking. The paper is structured as follows: Section 2 reviews the relevant literature, while Section 3 outlines the methodology. Section 4 describes the experimental setup and presents the results. Finally, Section 5 concludes the study by summarizing the findings, discussing limitations, and proposing future research directions.

## 2  Related Work

At least two streams of literature are relevant to the problem we are investigating: Explainable Artificial Intelligence (XAI) and adversarial attacks. Below, we discuss related works in each of these fields.

Research in XAI aims to enhance the understandability and interpretability of deep learning models, thereby increasing trust in artificial intelligence systems (Hassija et al., 2024). This field examines the extent to which inputs affect changes in outputs. In this paper, we concentrate on post hoc explanation methods. These methods explain how deep learning models decisions after they have been trained (Lopardo et al., 2024). Many of these post hoc explanation methods are gradient-based approaches. A simple example of gradient-based explanation involves analyzing the gradient of the model's output with respect to its input (Simonyan et al., 2014). Noteworthy gradient-based techniques include Gradient-weighted Class Activation Mapping (Grad-CAM) (Selvaraju et al., 2016), Layer-wise Relevance Propagation (LRP) (Bach et al., 2025), DeepLIFT (Deep Learning Important Features) (Li et al., 2021), Input Gradients (Hechtlinger, 2016), and SmoothGrad (Smilkov et al., 2017). Additionally, some post hoc explanation methods are based on perturbation. These approaches typically involve introducing perturbation to the input sample and observing how the output varies in response to these modifications. Among the popular perturbation-based methods are SHapley Additive exPlanations (SHAP) (Lundberg et al., 2017) and Local Interpretable Model-agnostic Explanations (LIME) (Ribeiro et al., 2016) both seek to measure the impact of input changes on the output, providing insights into the model's decision-making process.

On the other hand, research on adversarial attacks takes advantage of the "black-box" nature of deep learning models to exploit their vulnerabilities. These attacks involve creating inputs that are visually indistinguishable from humans but can deceive deep learning models such as the Fast Gradient Sign Method (FGSM) (Goodfellow et al., 2014) and the Basic Iterative Method (BIM) (Kurakin et al., 2018). Moreover, adversarial attacks can compromise model explanations. For instance, Ghorbani et al. (2019) showed that small, random perturbations to input data can significantly disrupt feature importance methods, such as saliency maps and DeepLIFT. Building on this, numerous studies have examined the impact of adversarial attacks on model explanations. These investigations employ various attack strategies, including adding perturbations to inputs (Kindermans et al., 2019; Dombrowski et al., 2019), altering model weights (Heo et al., 2019; Dimanov et al., 2020), and manipulating both training data and models to carry out more complex backdoor attacks (Viering et al., 2019; Noppel et al., 2023).

Despite significant research on adversarial attacks and explainable AI techniques, there remains a gap in the literature regarding a comprehensive benchmark for evaluating adversarial attacks specifically targeting explanation methods. Liu et al. (2021) developed the XAI-BENCH library, a benchmark focused on explanation techniques such as LIME and SHAP. OpenXAI, a benchmark aimed at post hoc explanation methods, was introduced by Agarwal et al. (2022). Yuan et al. (2019) provided a detailed project covering a wide range of adversarial attacks and their defense mechanisms, detailing the operating principles of advanced attacks and defenses. Baniecki and Biecek (2024) conducted a literature review on a broad spectrum of studies addressing adversarial attacks on explanation models, including various defense strategies. However, to the best of our knowledge, there is still no established benchmark for adversarial attacks on model explanations. Our goal is to fill this gap by compiling the results and establishing a benchmark.

## 3     Research Methodology

In this paper, our objective is to systematically compare the explanations of original images with those of adversarial images and to tabulate how these explanations alter when faced with adversarial manipulation. We use the ground truth explanation as a baseline control. First, we evaluate the explanations generated for original images against this control. Next, we perform a similar analysis for adversarial images, evaluating their explanations in relation to the control. Finally, we compare the results obtained from these two evaluations. Figure 3 outlines the steps involved in the methodology employed.

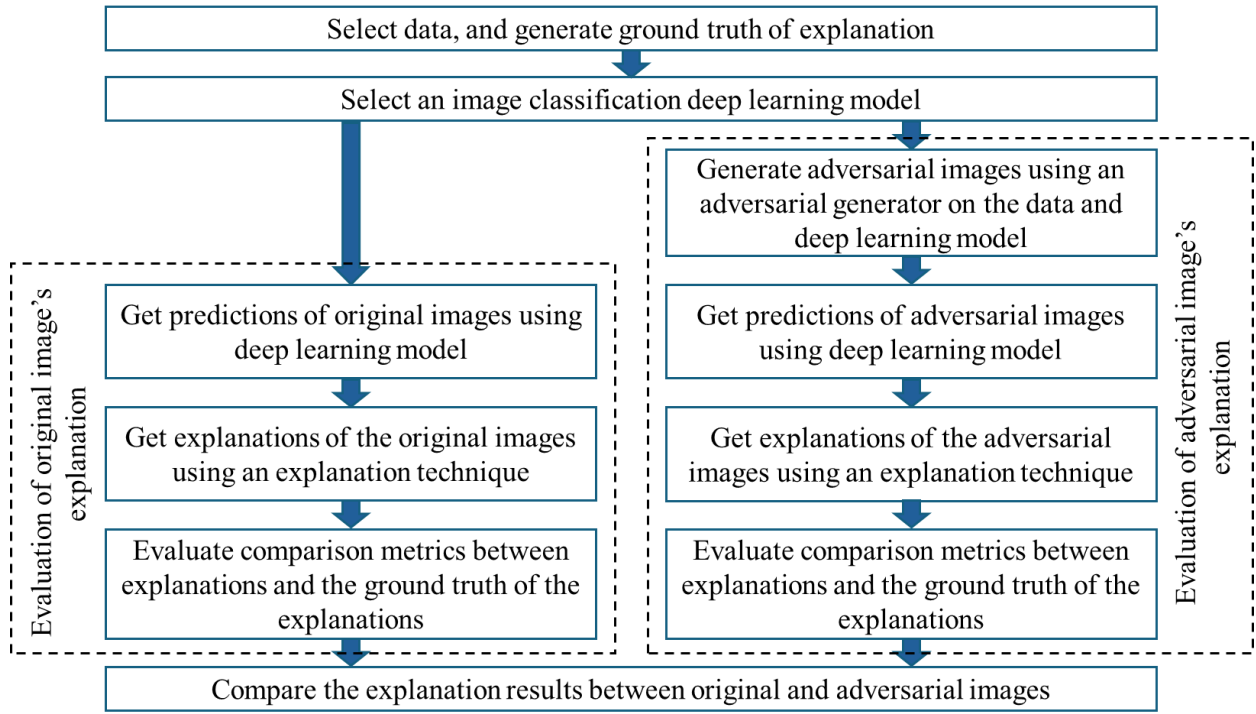

Figure 3: Outline of the research methodology

**Select data and generate ground truth of explanation:**

In this paper, we require image datasets accompanied by their corresponding ground truth explanations. In this context, by ground truth explanations, we imply an annotated part of the image responsible for its classification into a specific category. These annotations can be created by humans or through alternative methods. Please refer to Figure 4 to see sample images of the ground truth of the explanation.

In this study, we work with a specific subset of the ImageNet dataset (Deng et al., 2009), which encompasses 1,000 classes. Our chosen subset, sourced from Na (n.d.), consists of 5 images per class from the ImageNet validation dataset. We have selected images from classes 0 through 40 and 80 through 120 for our analysis, intentionally excluding classes 41 through 79. This exclusion was due to the presence of snake and lizard images in these classes, which we found to be emotionally depressing to handle manually.

Now, we are required to generate the ground truth of explanations of these images. We have used the Segment Anything Model (Kirillov et al., 2023) to develop the ground truth. SAM is itself a deep learning model designed for image segmentation. It combines convolutional neural networks with the attention mechanism (Vaswani et al., 2017) to identify and delineate boundaries of various objects within an image. SAM is capable of segmenting almost all types of objects within a picture.

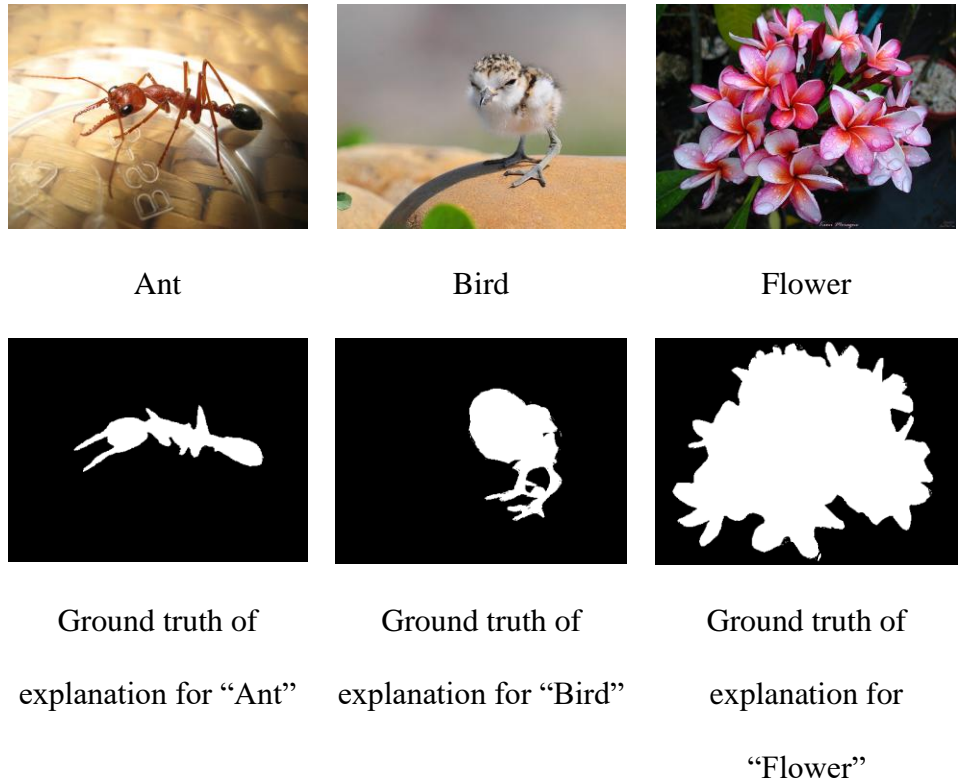

| Ant | Bird | Flower |

| Ground truth of explanation for "Ant" | Ground truth of explanation for "Bird" | Ground truth of explanation for "Flower" |

Figure 4: Sample images and their ground truth of explanations (Mohseni et al., 2021)

To generate masks using the SAM, we adopted a specific approach to ensure proper segmentation outcomes. We excluded images that contained multiple instances of objects to simplify the segmentation process. Additionally, we discarded grayscale images and retained only those images for which SAM successfully generated clear and accurate masks. This selective (image and mask selection) approach, done manually, helped us to generate effective and accurate ground truth for explanation.

**Select an image classification deep learning model:**

Next, we need to choose a deep learning model capable of classifying (image classification) the datasets. To achieve this, we have the option to develop a model from the ground up or leverage pre-trained models available within the TensorFlow (Abadi et al., 2016) library.

In our study, we utilized the pre-trained EfficientNetV2B0 (Tan & Le, 2021) model from the TensorFlow library, as described by Tan and Le in 2021. This model achieves a Top-1 accuracy of 78.7% (Keras, n.d.). on the ImageNet validation dataset. Top-1 accuracy refers to the model correctly identifying the true class as its top prediction. EfficientNetV2B0 incorporates several advanced features, such as depth-wise convolution and skip connections. For a detailed understanding of the model, we refer to Tan & Le (2021). The model comprises a total of 7.2 million parameters, with 7.13 million being trainable, striking a balance between computational demands and efficiency. Moreover, we found that explainable AI techniques were more effective with this model compared to other pre-trained models. All of these led us to choose this model for this study.

**Evaluations of original image's explanation:**

**Get predictions of the original images using deep learning model:**

Once the deep learning model and dataset have been selected, we can proceed to classify the images into their respective categories using the chosen deep learning model (with/without defense).

**Get explanations of the original images using an explanation technique:**

In the deep learning literature, there are quite a few techniques to explain why an image is classified into a particular class. Of these techniques, we employ GradCAM (Selvaraju et al., 2016), SmoothGrad (Smilkov et al., 2017), and LIME (Ribeiro et al., 2016) separately. We expect these explanation techniques to produce images similar to the ground truth shown in Figure 4. Now, we explain the key concepts behind GradCAM, SmoothGrad, and LIME.

GradCAM: GradCAM is a gradient-based explanation technique developed by Selvaraju et al. (2016). This technique is designed to explain the output of CNN models. CNN models can capture spatial information from images, and GradCAM utilizes this feature to identify which pixels are most influential in determining the model's output. GradCAM computes the gradient of the target class score (the output before activation) with respect to the feature maps of a convolutional layer. This expresses how each pixel affects the class score. Subsequently, Global Average Pooling (GAP) is applied to these gradients to generate a set of weights for each feature map. These weights are then used to create a weighted combination of the feature maps, producing a heatmap. When this heatmap is overlaid on the input image, we can see which areas in the image are responsible for the model's prediction.

SmoothGrad: SmoothGrad, developed by Smilkov et al. (2017), is a gradient-based technique designed to refine the interpretability of neural network decisions by producing smoother saliency maps. The method involves adding random noise to create multiple copies of an original image and computing a saliency map for each noisy version; these maps depict how pixel variations affect the model's output. By averaging these saliency maps, SmoothGrad reduces the noise typically found in individual maps. This technique expresses which features a neural network considers most important during calculating an output.

LIME: LIME is a model-agnostic explanation technique developed by Ribeiro et al. (2016). This technique begins by generating perturbed samples, which are created by introducing noise to the original image. These samples are then weighted based on their similarity to the original image, with samples that are more similar receiving higher weights. Subsequently, a simple model, such as linear regression, is trained on these weighted samples. The coefficients of this simple model indicate the influence of each input feature on the output, thus revealing which features are most responsible for the model's prediction.

**Evaluate comparison metrics between explanations and the ground truth of the explanations:**

Next, we compare the generated explanations with the ground truth explanations. Various metrics can be utilized for this comparison; however, for this study, we use root mean square error (RMSE) and intersection over union (IoU). These two metrics are defined mathematically as follows:

If $Y$ and $\hat{Y}$ represents the ground truth and prediction explanation of an image, then RMSE and IoU can be described as

$$\text{RMSE} = \sqrt{\frac{1}{n^2}(\sum_i^n \sum_j^n (Y_{ij} - \hat{Y}_{ij})^2)}$$

$$\text{IoU} = \frac{pixels(Y \cap \hat{Y})}{pixels(Y \cup \hat{Y})}$$

**Evaluations of adversarial images's explanation:**

**Generate adversarial images using an adversarial generator on the data and deep learning model:**

In the deep learning literature, there are two primary categories of adversarial attacks: targeted and non-targeted. In a targeted attack, the goal is to manipulate a model into classifying an adversarial image as a specific, incorrect label chosen by the attacker. Conversely, a non-targeted attack aims to cause the model to incorrectly classify the adversarial image as any label other than the true one. We use the non-targeted attack methodology to generate the adversarial images.

We can find a wide range of methods for generating non-targeted adversarial images. However, we use the FGSM (Goodfellow et al., 2014) and the BIM (Kurakin et al., 2018) for generating adversarial images in this study. The key concepts of these adversarial generation methods are described below:

FGSM: Initially, it was believed that the vulnerability of deep learning models to adversarial attacks stemmed from the deep learning model's nonlinearity and tendency to overfit. However, Goodfellow et al. (2014) suggested that the real issue lies in their inherent linearity and capacity for generalization. This insight led to the development of the Fast Gradient Sign Method (FGSM), a simple yet effective technique applicable to both targeted and non-targeted attacks. FGSM works by introducing slight perturbations in the original image, aligned with the gradient of the loss function relative to the model's output. The underlying equation for this is as below:

$$x_{adv} = x + \epsilon \, \text{sign}(\nabla_x J(\theta, x, y))$$

In this equation, $x_{adv}$ denotes the adversarial image generated by adding a small perturbation to the original sample $x$. The model parameter is expressed by $\theta$ and the correct label of the original image is $y$. The term $\nabla_x J(\theta, x, y)$ captures the gradient of the loss function with respect to $x$, and $\epsilon$ is a small value ensuring the adversarial image remains visually indistinguishable to humans.

BIM: This method is built upon FGSM. Kurakin et al., (2018) introduced two iterative methods to enhance attack success rates. Among these, the Basic Iterative Method applies the FGSM multiple times with pixel values clipped to ensure they remain within a specific range of the original image, resulting in a more effective adversarial attack. The iterative process follows these equations (Kurakin et al., 2018):

$$X_0^{adv} = X$$

$$X_{N+1}^{adv} = Clip_{X,\epsilon}\{X_N^{adv} + \alpha \text{sign}\left(\nabla_x J(X_N^{adv}, y_{true})\right)\}$$

This method starts with the original image as the initial adversarial sample $X_0^{adv}$. It then iteratively adds small perturbations aligned with the loss gradient, clipping the result after each step to ensure the adversarial images remain close to the original images.

**Get predictions of the adversarial images using the deep learning model:**

We employ the same deep learning model that was used for classifying the original images to also classify the adversarial images.

**Get explanations of the adversarial images using an explanation technique:**

We apply the same explanation technique used for the original images to generate explanations for the predictions of the adversarial images.

**Evaluate comparison metrics between explanations and the ground truth of the explanations:**

We use the same metrics as before to compare the generated explanations with the ground truth explanations.

**Compare the explanation results between original and adversarial images:**

Finally, we compare original and adversarial images based on the metrics obtained from evaluating explainable AI techniques against the ground truth. By doing so, we get an idea of the adversarial impact on explainable techniques.

We designed the experiment to include a baseline control measure, addressing the potential inaccuracies in prediction explanations from original images. By employing baseline control, we ensure a more scientific approach in comparing explanations between original and adversarial images, thus establishing a more rigorous evaluation framework. The results are systematically tabulated to identify any discernible patterns.

## 4    Experimental Platform

Dataset: We utilized a subset of the ImageNet validation dataset (Deng et al., 2009), specifically classes 1-40 and 80-120, as sourced from Na (n.d.). Each of these classes contains 5 images, though we did not use all five images from each class. Instead, we only selected images from these classes for which the ground truth of the explanation was accurately generated by SAM.

Deep learning model: We have used Pre-trained EfficientNetV2B0 (Tan & Le, 2021). This model achieves a Top-1 accuracy of 78.7% (Keras, n.d.). on the ImageNet validation dataset. Top-1 accuracy refers to the model correctly identifying the true class as its top prediction. The model comprises a total of 7.2 million parameters, with 7.13 million being trainable, striking a balance between computational demands and efficiency. The accuracy of the deep learning image classification model on our subset of original images is 89.94%.

Comparison metric: We have used RMSE and IoU to compare the generated masks with the ground truth.

Software and hardware: We conducted our experiments using Python in the Google Colab Pro environment, equipped with a high RAM (25 GB) configuration and a T4 GPU.

## 5　Results and Discussion

We present our results step by step. Initially, we display sample images alongside their ground truth generated by SAM. Following this, we will show the masks created by various explanation techniques. Subsequently, we demonstrate how adversarial images affect model performance while remaining indistinguishable from human eyes. We then present the masks derived from these adversarial images. Afterward, we compare these masks with the ground truth. Finally, we compile and tabulate the results of these comparisons.

In Figure 5, we display sample images from the dataset alongside their corresponding ground truths of explanation, which were generated by SAM through a careful manual selection process. In Figure 6, we show sample images along with explanations obtained from various explanation techniques. It is important to note that these explanations were generated by selecting the top 15% of pixels that contribute to the classification.

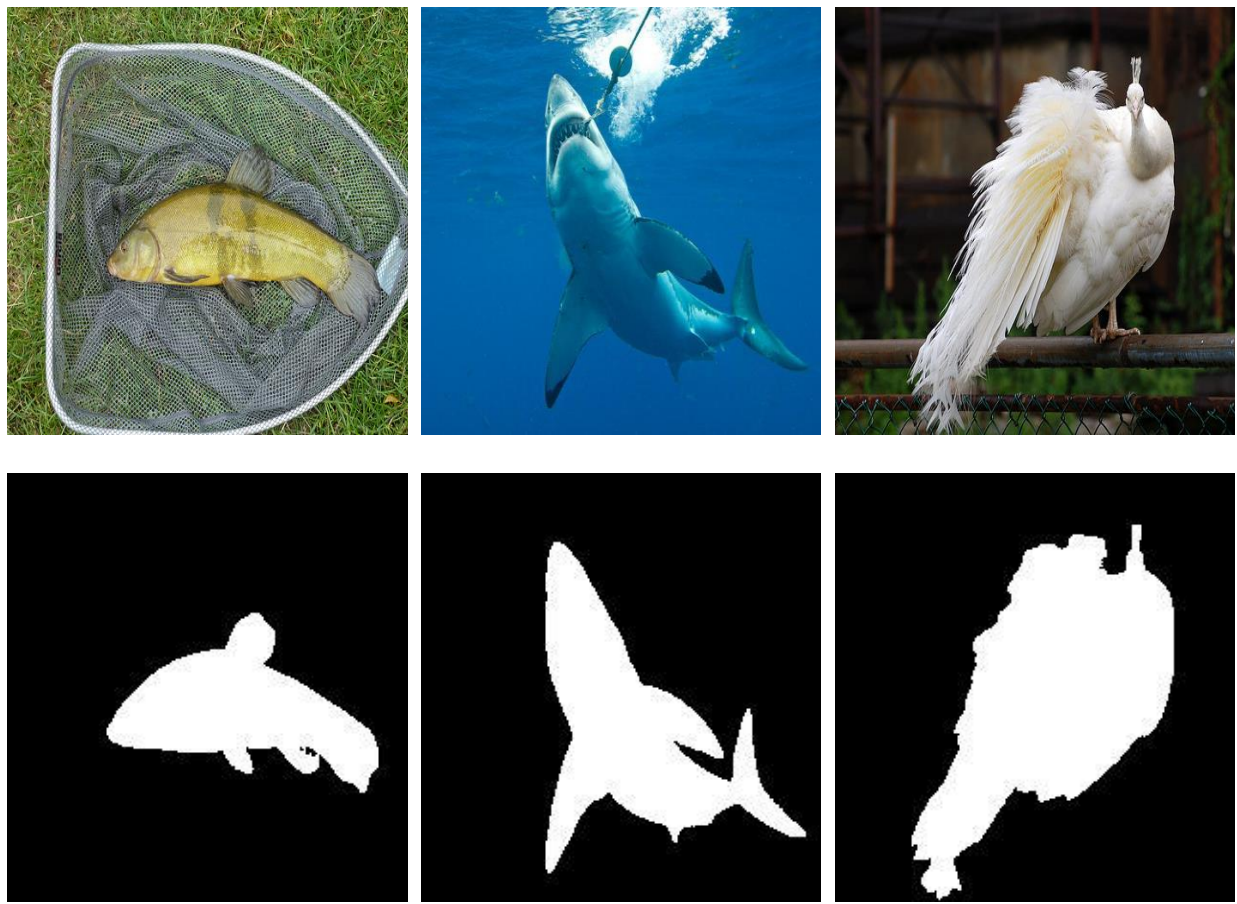

(a) Class 0: Tinca tinca      (b) Class 2: White shark      (c) Class 84: Peacock

Figure 5: Sample images in the dataset with their ground truth of the explanation

In Figure 7, we present sample images along with their adversarially modified images generated using the FGSM and BIM methods. Visually, these adversarial images are indistinguishable from the originals. However, their impact on model accuracy is significant. Without any attack, the accuracy of our deep learning model is 89.94%. After employing the FGSM attack with an epsilon value of 2.5%, accuracy drops to 58.73%. The impact is even greater with the BIM attack at the same epsilon value over 10 iterations, reducing accuracy to 45.50%. Despite the visual similarities, these adversarial attacks significantly degrade model performance.

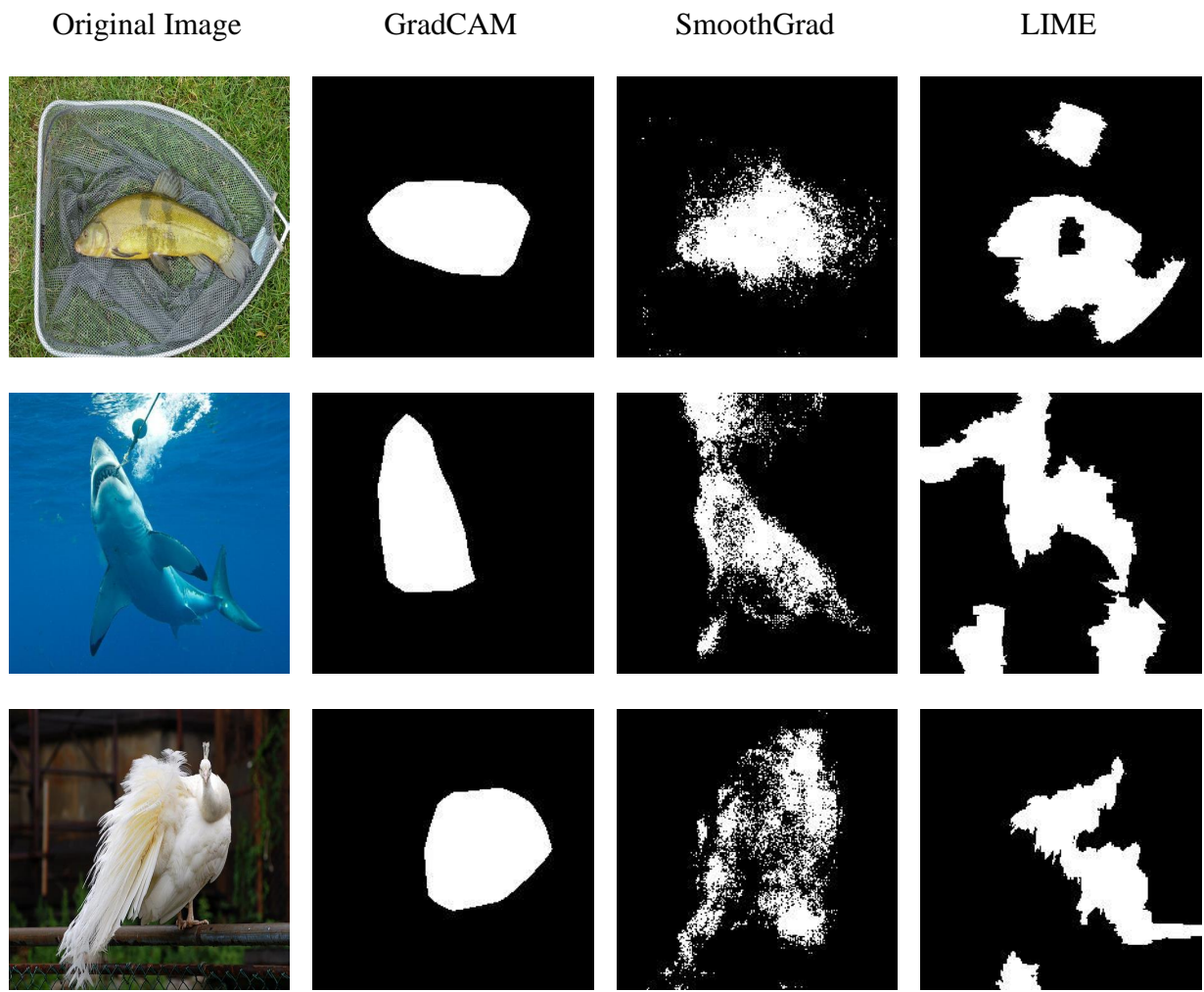

Figure 6: Sample images and their explanation from different explanation techniques

Figure 8 presents sample adversarial images created using the FGSM attack along with explanations from various explanation techniques. Similarly, Figure 9 shows adversarial images generated by the BIM attack, accompanied by explanations from different explanation techniques. In these figures, we observe how the explanations change following the adversarial attacks.

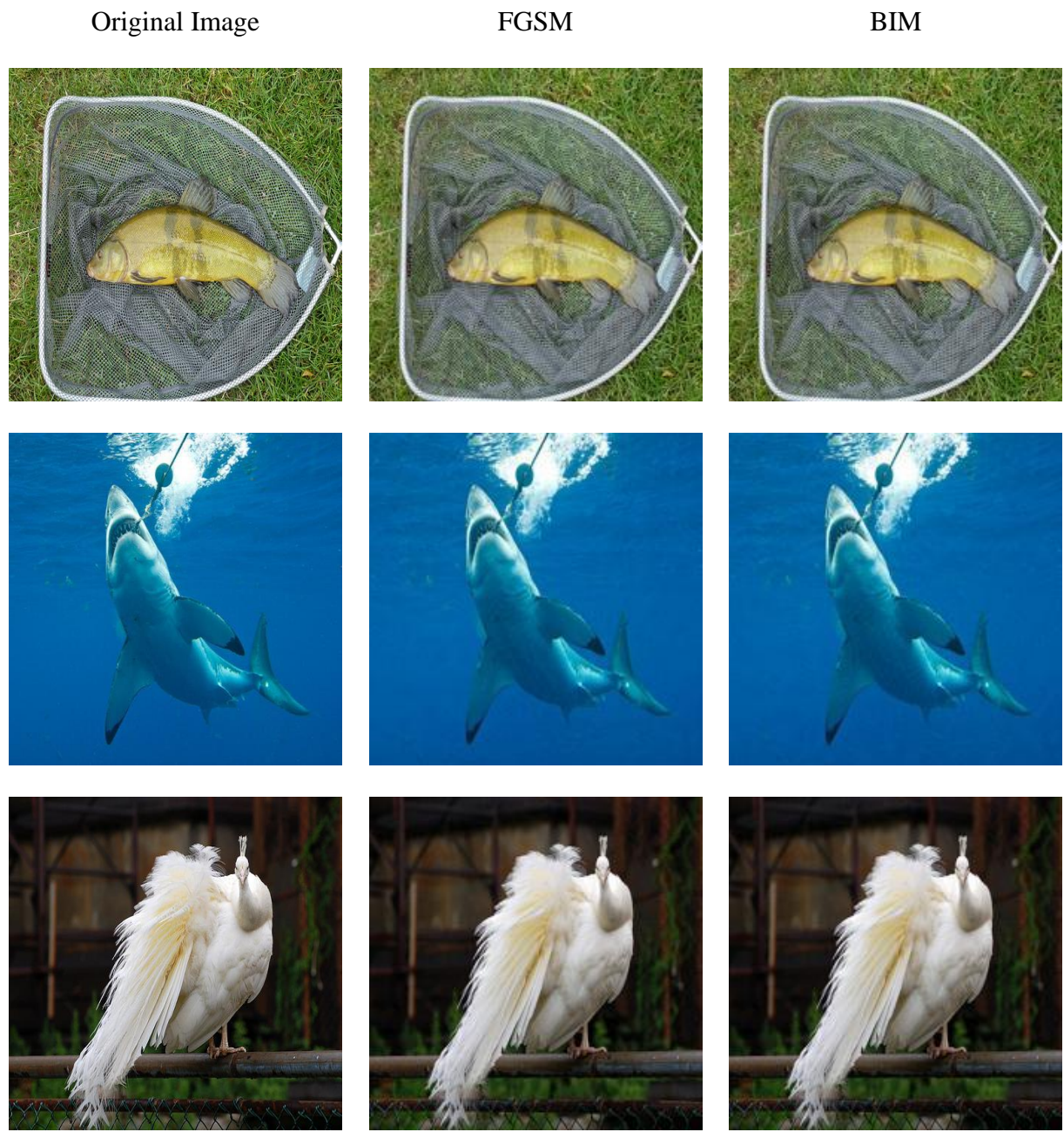

Figure 7: Adversarial image generation using FGSM and BIM

| Avderarial Image (FGSM) | GradCAM | SmoothGrad | LIME |

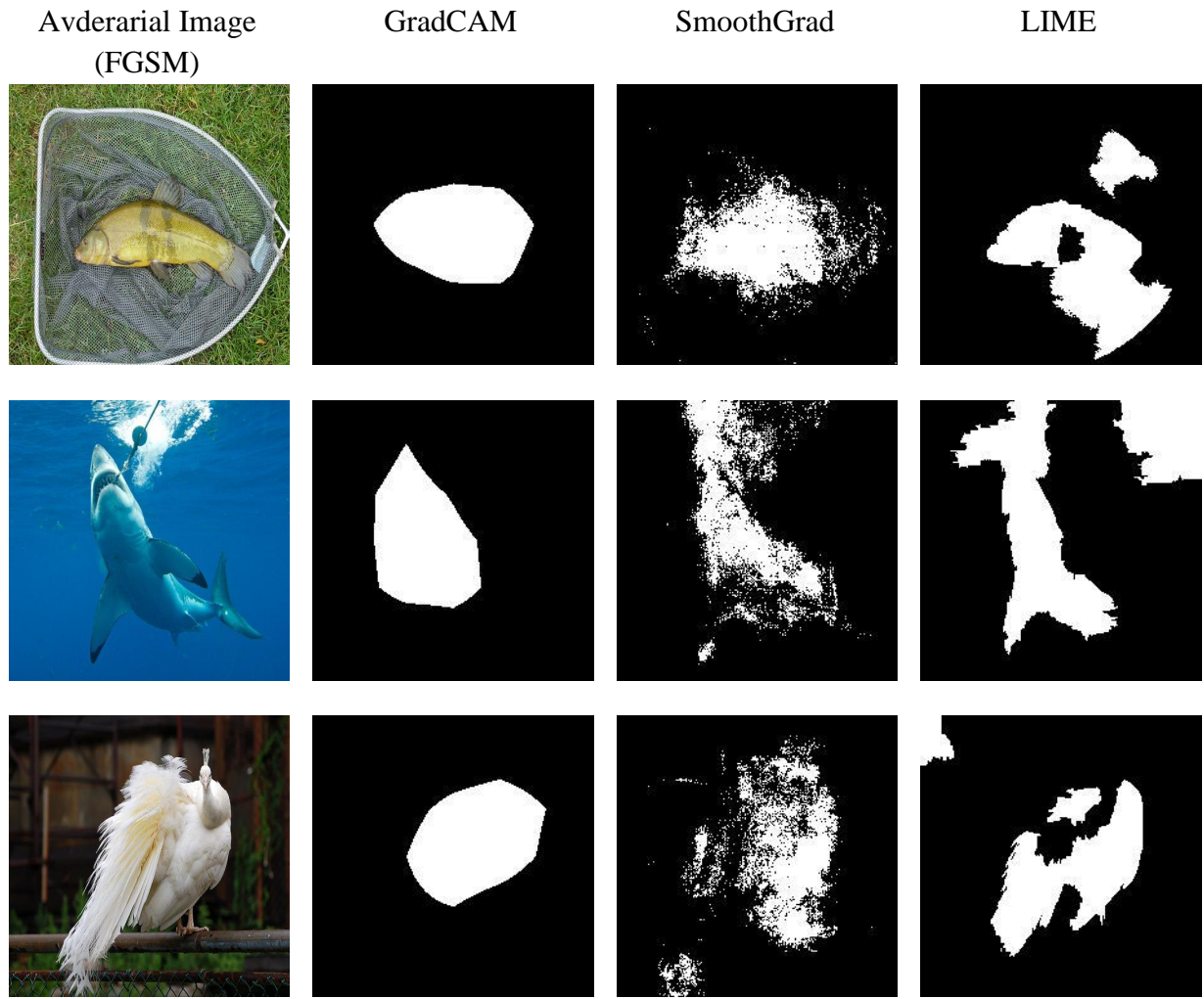

Figure 8: Sample adversarial images (FGSM) and their explanation from different explanation techniques

For this study, we analyzed a total of 189 images, applying the methodology outlined in the study to each image individually. We then aggregated the results to establish a benchmark that demonstrates how various model explanations are influenced by adversarial attacks. We assessed the explanation masks produced by different techniques against the ground truth using the Intersection over Union (IoU) and Root Mean Square Error (RMSE) metrics.

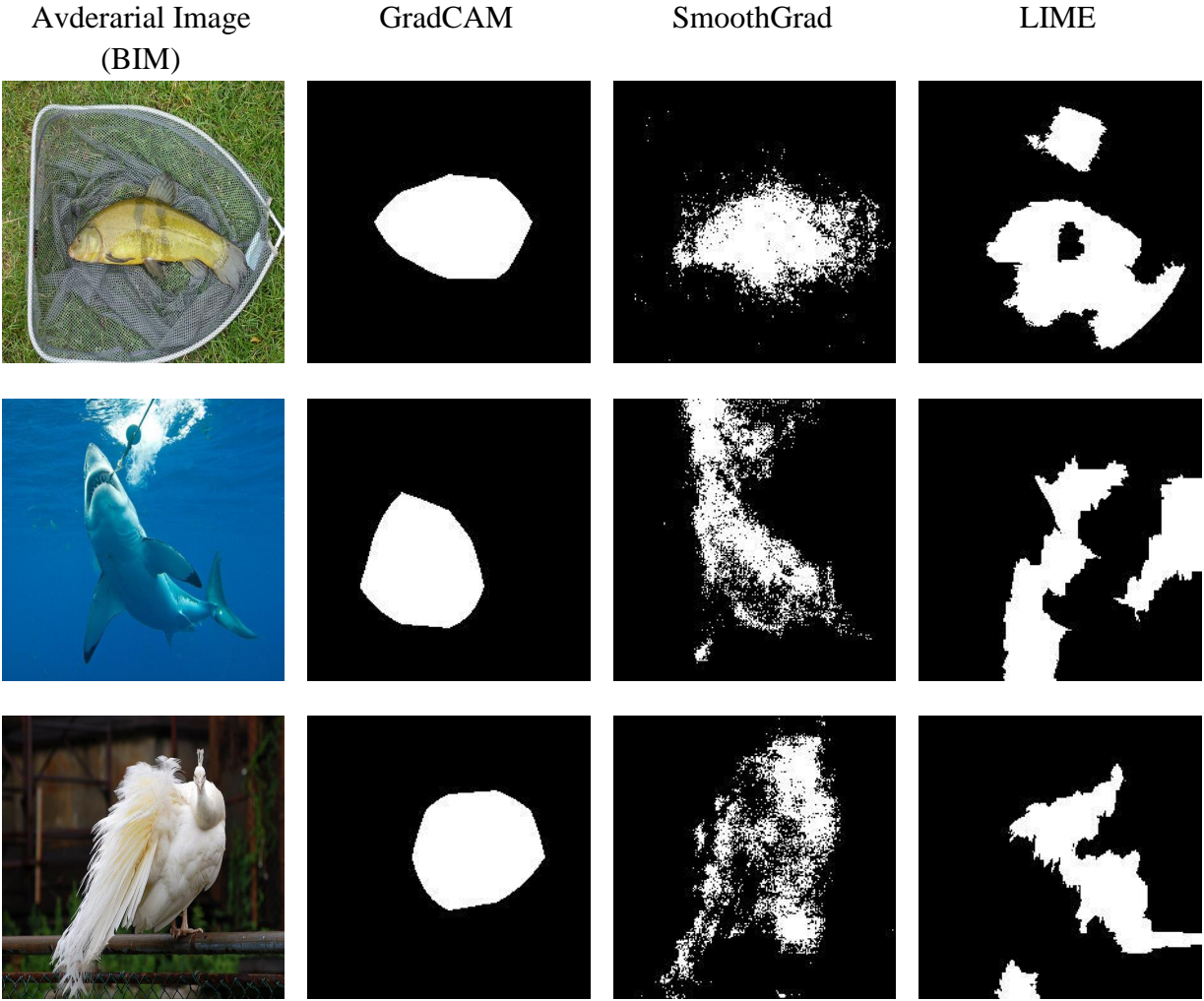

Figure 9: Sample adversarial images (BIM) and their explanation from different explanation techniques

Table 1 displays the IoU values for different explanation techniques, both with and without adversarial attacks. A higher IoU value indicates that the explanation technique more effectively identifies crucial image regions. Our findings show that GradCAM consistently achieves the highest IoU in all scenarios, whereas SmoothGrad shows poorer performance.

Similarly, Table 2 presents the RMSE values for various explanation techniques under the same conditions. From the results, we find SmoothGrad consistently achieves the lowest RMSE, indicating high accuracy in its explanations. It is important to note that the RMSE and IoU metrics show minimal variation across different explanation techniques, regardless of whether they are subjected to adversarial attacks. However, adversarial attacks lead to a substantial decrease in the test accuracy of the deep learning model, dropping from 89% to approximately 50%.

Table 1: Changes in explanation compared to ground truth using IoU

|  | Explanation Methods | | |
| --- | --- | --- | --- |
|  | SmoothGrad | GRADCAM | LIME |
| Without Attack | 19.83% | 34.66% | 24.78% |
| FGSM-2.5% | 19.63% | 33.67% | 24.57% |
| BIM-2.5% | 19.46% | 32.75% | 24.63% |

Table 2: Changes in explanation compared to ground truth using RMSE

|  | Explanation Methods | | |
| --- | --- | --- | --- |
|  | SmoothGrad | GRADCAM | LIME |
| Without Attack | 36.21% | 37.40% | 43.28% |
| FGSM-2.5% | 36.46% | 37.95% | 43.59% |
| BIM-2.5% | 36.66% | 38.43% | 44.37% |

# 6    Conclusion

In this paper, we started the study by selecting a dataset and generating a ground truth explanation using the SAM. Next, we employed GradCAM, SmoothGrad, and LIME techniques to find deep-learning model explanations for the original images. We then created adversarial images using the FGSM and BIM methods and obtained explanations for these adversarially generated images as well. The final step involved calculating the IoU and RMSE to compare these explanations against the ground truth.

The primary aim of this paper was to explore how the explanations of a deep learning model for an image classification task are changed in the presence of adversarial attacks and to assess the effectiveness of different explanation techniques under adversarial attacks. We observed a significant drop in the test accuracy of the deep learning model, from 89.94% to 58.73% following an FGSM attack, and to 45.50% after a BIM attack. However, we noted no significant changes in the IoU and RMSE metrics for the explanations after the attacks, suggesting that these explanation methods with the combination of IoU and RMSE metrics are not effective in discerning adversarial influences.

Regarding limitations, we acknowledge that we did not incorporate a comprehensive range of adversarial generation methods, metrics, and explanations to establish a complete benchmark. Additionally, the analysis was conducted on a very small subset of data, which restricts the generalizability of the findings. For future directions, these limitations can be addressed by including a broader variety of adversarial attack methods, metrics, and explanation techniques, as well as by expanding the dataset used for analysis. Furthermore, benchmarking the impact of implementing various adversarial defense mechanisms presents another promising direction.

The necessary programming codes, along with images, can be found in the following repository: https://github.com/ahnafsadat/Explainable_AI_evaluation

Dombrowski, A. K., Alber, M., Anders, C., Ackermann, M., Müller, K. R., & Kessel, P. (2019). Explanations can be manipulated and geometry is to blame. *Advances in neural information processing systems*, *32*.

Ghorbani, A., Abid, A., & Zou, J. (2019, July). Interpretation of neural networks is fragile. In *Proceedings of the AAAI conference on artificial intelligence* (Vol. 33, No. 01, pp. 3681-3688).

Goodfellow, I. J., Shlens, J., & Szegedy, C. (2014). Explaining and harnessing adversarial examples. *arXiv preprint arXiv:1412.6572*.

Hassija, V., Chamola, V., Mahapatra, A., Singal, A., Goel, D., Huang, K., ... & Hussain, A. (2024). Interpreting black-box models: a review on explainable artificial intelligence. *Cognitive Computation*, *16*(1), 45-74.

Hechtlinger, Y. (2016). *Interpretation of Prediction Models Using the Input Gradient*. http://arxiv.org/abs/1611.07634

Heo, J., Joo, S., & Moon, T. (2019). Fooling neural network interpretations via adversarial model manipulation. *Advances in neural information processing systems*, *32*. https://github.com/SinaMohseni/ML-Interpretability-Evaluation-Benchmark

Keras. (n.d.). Keras Documentation: Keras applications. https://keras.io/api/applications/

Khattar, A., & Quadri, S. M. K. (2022). Generalization of convolutional network to domain adaptation network for classification of disaster images on twitter. *Multimedia Tools and Applications*, *81*(21), 30437-30464.

Kindermans, P. J., Hooker, S., Adebayo, J., Alber, M., Schütt, K. T., Dähne, S., ... & Kim, B. (2019). The (un) reliability of saliency methods. *Explainable AI: Interpreting, explaining and visualizing deep learning*, 267-280.
26